\documentclass[11pt,a4paper]{article}
\usepackage[hyperref]{emnlp2020}
\usepackage{times}
\usepackage{latexsym}

\usepackage{microtype}

\usepackage{microtype}
\usepackage{amsmath}
\usepackage{cleveref}
\usepackage{graphicx}
\usepackage{amsthm}
\usepackage{multirow}
\usepackage{subfigure}
\usepackage{booktabs}
\usepackage{amssymb}
\usepackage{url}
\usepackage{amsmath}

\usepackage{url}
\usepackage{cleveref}

\crefformat{section}{\S#2#1#3} 
\crefformat{subsection}{\S#2#1#3}
\crefformat{subsubsection}{\S#2#1#3}

\definecolor{skyblue}{HTML}{2554C7} 

\aclfinalcopy 

\title{Multilevel Text Alignment with Cross-Document Attention}

\author{
Xuhui Zhou$^{\heartsuit}$\quad  
Nikolaos Pappas$^{\clubsuit}$\quad 
Noah A. Smith$^{\clubsuit}$$^{\diamondsuit}$\\
\\
$^{\heartsuit}$Department of Linguistics, University of Washington \\
  $^{\clubsuit}$Paul G.\ Allen School of Computer Science \& Engineering, University of Washington \\
   $^{\diamondsuit}$Allen Institute for Artificial Intelligence\\
   {\tt xuhuizh@uw.edu,\{npappas,nasmith\}@cs.washington.edu } 
}

\date{}

\begin{document}
\maketitle
\begin{abstract}
Text alignment finds application in tasks such as citation recommendation and plagiarism detection. Existing alignment methods operate at a single, predefined level and cannot learn to align texts at, for example, sentence \emph{and} document levels. We propose a new learning approach that equips previously established hierarchical attention encoders for representing documents with a cross-document attention component, enabling structural comparisons across different levels (document-to-document and sentence-to-document). Our component is weakly supervised from document pairs and can align at multiple levels. Our evaluation on predicting document-to-document relationships and sentence-to-document relationships on the tasks of citation recommendation and plagiarism detection shows that our approach outperforms previously established hierarchical, attention encoders based on recurrent and transformer contextualization that are unaware of structural correspondence between documents.  
\end{abstract}

\section{Introduction}

Aligning texts and understanding their relationships is a common problem for NLP tasks such as citation recommendation \cite{Bhagavatula2018ContentBasedCR,jiang2019}, comparable document mining \cite{he10,peng-etal-2016-news,Bhagavatula2018ContentBasedCR,guo-etal-2019}, parallel sentence mining \cite{shi-etal-2006-dom,ture-lin-2012-grab,guo-etal-2018-effective}, plagiarism detection \cite{barron-cedeno-etal-2010-plagiarism,forner13,ferrero-etal-2017-using}, paraphrase identification \cite{wan-etal-2006-using,das-smith-2009-paraphrase,wang-etal-2016-sentence}, and  textual entailment \citep{Dagan2004PROBABILISTICTE,androutsopoulos10,zhao-etal-2016-textual}. Longer texts make the problem more challenging due to the potential complexity of the underlying correspondence.\begin{figure}[t]
\centering 
\includegraphics[width=\columnwidth]{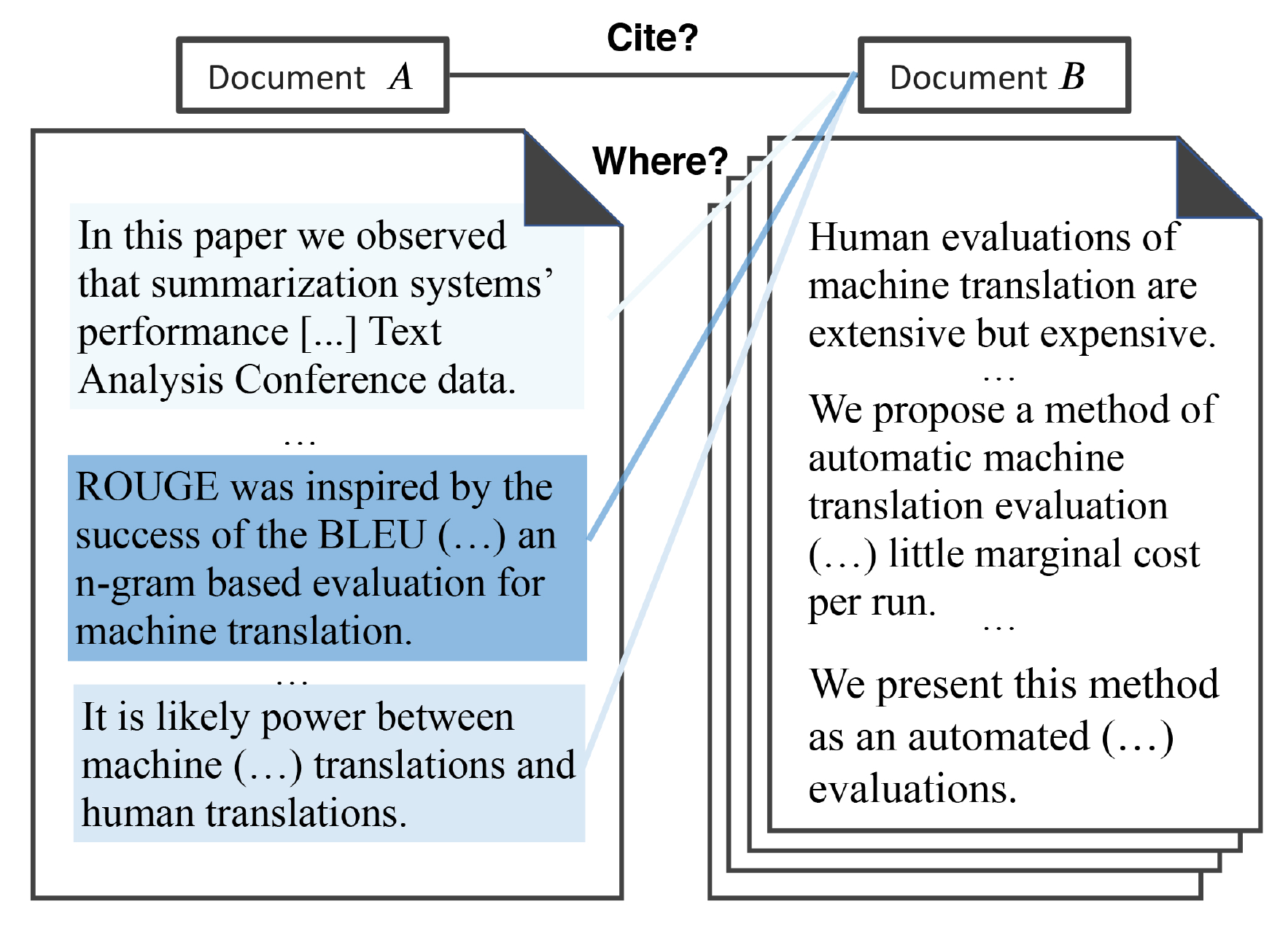} 
\caption{A motivating example of aligning scientific documents at different levels.  We consider citation recommendation (whether $A$ cites $B$) and citation localization (which sentence in $A$ cites) at the same time. The confidence of our model for citation  localization is represented by the degree of blueness.
\label{fig:example}}
\end{figure}
 Here, we develop a model to address this problem and demonstrate its applicability on three different tasks which require such understanding, namely on citation recommendation, citation localization, and plagiarism detection for general web documents. 
 
 One key component of an NLP system for aligning documents is the encoding process. Present approaches for comparing documents rely on hierarchically structured document encoders such as hierarchical attention networks (HANs; \citealp{yang16}), which  independently represent the two documents as fixed-length vectors. The vectors are fed to a classifier which makes a decision on the relation between them \cite{jiang2019,guo-etal-2019}. 
 However, such methods do not provide insights about or leverage the underlying relationships across documents and are applicable only to a single, predefined level \cite{jiang2019, yang2020512}. Importantly, when comparing documents, those methods ignore the structural correspondence between parts. 
 
Figure \ref{fig:example}  shows an example of predicting and localizing citations in scientific documents. To solve this problem in a cost-effective way, models need to be able to make joint predictions about these different tasks without relying on fine-grained annotations which are typically more expensive to obtain. In this paper, we propose a new approach for encoding documents that aligns document parts during the encoding process and is able to make predictions about their relationships across different levels (specifically, document-to-document and sentence-to-document). In particular, we equip a powerful class of models, namely hierarchical attention encoders \cite{yang16,liu-lapata-2019,guo-etal-2019} with a cross-document attention component that ``attends'' to the structure of documents, enabling inferences about alignment of their parts (Section~\ref{sec:approach}).

We introduce new benchmarks for joint document-to-document prediction and sentence-to-document localization of document relationships for citation recommendation and plagiarism detection (Section~\ref{sec:benchmark}).  Our experiments with variations find that cross-document attention is beneficial to strong baseline hierarchical encoders (Section~\ref{sec:experiments}) on these challenging tasks.

\section{Comparing Documents}
\label{sec:problem}

Many potential applications of natural language processing involve a comparative analysis of two (or more) documents.  Examples include:
\begin{itemize}
    \item recommending existing documents to be cited in a new document \citep{jiang2019, yang2020512};
    \item inferring whether one document plagiarizes another \citep{Foltnek2019AcademicPD};
    \item inferring whether one document is a translation of another \citep{guo-etal-2019}; and
    \item multi-document summarization \citep{liu-lapata-2019} and coreference resolution \citep{lee12}.
\end{itemize}
Our experiments in Section~\ref{sec:experiments} will consider tasks inspired by the first two applications.

Note that, in each of these examples, the most useful analysis of the document-to-document relationship will include a more fine-grained analysis:  which \emph{parts} of the source document correspond to which parts of the target document?  Figure~\ref{fig:example} illustrates an example for citation recommendation, in which 
the main relationship (does/should document $A$ cite document $B$?) is actually composed of a number of more localized relationships between sentences in document $A$ that contain citations and document ${B}$ (or, perhaps, parts of document $B$).  In general, whenever we seek to model relationships between documents, we believe that these relationships can be \emph{localized} in one or both documents.  We believe that automatic identification of these local correspondences is useful, both directly (e.g., mining parallel sentences for use in training a machine translation system), and for providing explanations (e.g., in plagiarism detection).

Of course, these fine-grained, localized correspondences are not typically directly observable in realistic datasets.  Here, we consider scenarios where positive and negative examples of document-level relationships are available for supervision, but fine-grained correspondences between their parts are not.  We exploit simple decompositions of documents (into sentences and words) but follow earlier work \citep{yang16} in offering a general hierarchical model that could be extended to allow for additional levels in future work.

The problem we aim to solve is:  (i) given two documents (each decomposed, e.g., into sentences and words), automatically categorize whether a particular relationship holds between them, and (ii) which parts between them should be ``aligned'' in support of the relationship in (i).  We will refer to these tasks respectively as document-to-document alignment (D2D) and sentence-to-document alignment (S2D), 
and will conduct experiments on tasks of both kinds in Section~\ref{sec:experiments}.

\section{Approach}
\label{sec:approach}

We next describe our solution to this problem, starting with a high-level overview (Section~\ref{sec:overview}).
We build on a family of widely used models for document representation, known as hierarchical attention networks (HANs; Section~\ref{sec:han}), which is sensitive to predefined notions of hierarchy (here, sentences; \citealp{yang16}).
We augment the HAN with cross-document attention  (Section~\ref{sec:cross-document}).

\subsection{Overview} \label{sec:overview}

The training data assumed in our setup is a collection of labeled document pairs.  In this work, the labels are binary (either the relationship of interest exists or does not).  Let $\langle A, B, y\rangle$ denote a training tuple of two documents with their label $y$.
We apply a familiar ``Siamese'' architecture \cite{Mueller2016SiameseRA,jiang2019}:  $A$ and $B$ are encoded using the same function (which we call the ``document encoder''), the outputs are concatenated, and then passed through a fully-connected relu layer and a sigmoid function to yield a score.  The network is trained to minimize cross-entropy.

This model relies heavily on the document encoder to learn representations relevant to the relationship of interest.   As discussed in Section~\ref{sec:problem}, we desire an encoder that can align parts of either or both texts, localizing the relationship to particular sentences, but any encoding function for a document can be used.  Our baselines, based on the encoder we present next, do not have any notion of alignment, while our new model does (Section~\ref{sec:cross-document}).

\begin{figure*}
\centering
\includegraphics[width=0.97\textwidth]{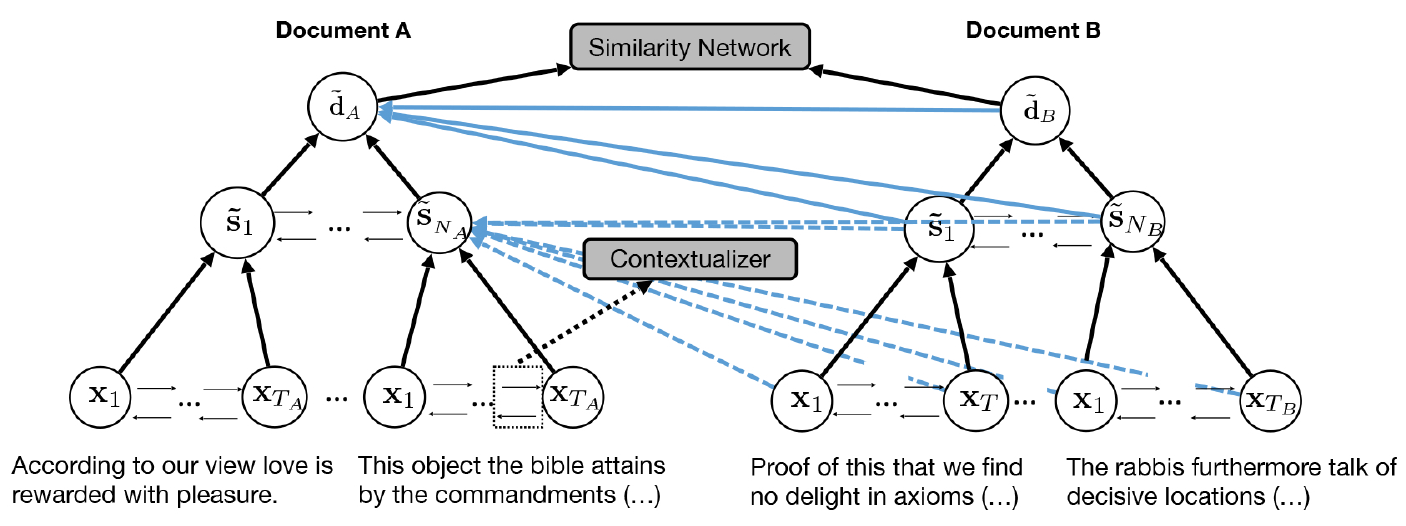} 
\vspace{-1mm}
\caption{\label{fig:model} Illustration of our models. In the exposition, \textsc{Shallow} considers cross-document attention without the dashed line, while \textsc{Deep} considers all levels' cross-document attention. Only part of attention is shown in the figure for clarity. 
The similarity network is for predicting the binary label. The  \textcolor{skyblue}{blue} arrows indicate the cross-document attention for two example nodes, namely $\tilde{\mathbf{d}}_A$ and $\tilde{\mathbf{s}}_N$. 
}
\end{figure*}

\subsection{Hierarchical Attention Networks} \label{sec:han}

\citet{yang16} introduced a family of document encoding models that are based on a word/sentence/document hierarchy, known as hierarchical attention networks (HANs).
They have been shown superior to earlier hierarchical encoding models based on convolutional networks \cite{collobert11,kim14,zhang15}, they are competitive for tasks involving long documents 
\citep{choi16,pappas17c,sun-etal-2018-stance,miculicich-etal-2018,liu-lapata-2019,guo-etal-2019},\footnote{Transformers have emerged as a successful tool across NLP \citep{NIPS2017_7181}, but they are not yet well suited for long sequences without an hierarchical configuration because their costs scale quadratically with sequence length.
When more efficient variants of transformers become available, they will be an appealing option to consider in this setting as well.} and they can be used orthogonally to other design decisions (e.g., word embeddings and the use of pretraining).

For document $X$, a HAN builds a vector representation $\mathbf{d}_X$ using the (given) structure of $X$:  typically, the document vector is derived from sentence vectors, which are derived from (contextualized) word vectors.  Working in the order that the computation proceeds, the encoding procedure is:
\begin{enumerate}
    \item Each word in the document is mapped (by lookup) to its type embedding.\footnote{``Type embedding'' refers to traditional word (subword) vectors;
    we use the term to contrast with contextualized embeddings associated with specific tokens.}
    \item Each word's vector  is contextualized, i.e., a new word \emph{token} vector is derived from the word and the other words in the sentence.  In this work, we consider two contextualizers:  pretrained BERT \citep{devlin19} and a GRU \citep{cho-etal-2014} 
    whose parameters are trained only for the end task.
    \item Each sentence in the document is encoded by aggregating the contextualized word vectors.  Letting $\mathbf{x}_i$ denote the $i$th word vector and $\mathbf{y}$ denote the sentence vector, the layer that performs this aggregation has the form:
    \begin{align}
        \mathbf{y} &= \sum_i \overbrace{\frac{ \exp \big[ \mathbf{u}^\top \tanh \big( \mathrm{affine}(\mathbf{x}_i) \big) \big]}{\sum_j \exp \big[ \mathbf{u}^\top \tanh  \big( \mathrm{affine}(\mathbf{x}_j) \big) \big]}}^{\textrm{attention}_i} \mathbf{x}_i, \label{eq:aggregation}
    \end{align}
    where $i$ and $j$ range over the words within the sentence.  We suppress the parameters of the affine transformation but not the attention parameters $\mathbf{u}$. Note that, when using pretrained BERT, we instead take the average of word token vectors to obtain sentence vectors, following \citet{reimers-gurevych-2019-sentence}.
    
    \item Analogous to the two steps above, the sentence vectors are contextualized using a bidirectional GRU for both word-level contextualizers (the default encoder for HAN at the sentence level; \citealp{yang16}) 
    and then aggregated.  Aggregation is exactly as in Equation~\ref{eq:aggregation}, but $\mathbf{x}_i$ now denotes a contextualized sentence vector and $\mathbf{y}$ is the document vector $\mathbf{d}$.  
    A separate set of parameters is used at this level of the hierarchical model. Note that the contextualization can be done with transformers too as by \citet{Pappagari_2019}; we leave this alternative option as future work.
\end{enumerate}

HANs handle long documents by imposing a simple notion of hierarchy and compositionality; they restrict the dependence of one part's representation on the representations of its neighbors.  They have been used effectively for semantic comparison tasks between documents \citep{jiang2019}, but they do not offer a way to localize correspondences between parts of the two documents.

\subsection{Cross-Document Attention} \label{sec:cross-document} 
We augment HANs with a cross-document attention (CDA) mechanism that attends to their document parts,  allowing them to reason over structural correspondences between documents. 
  Illustrated in Figure~\ref{fig:model}, the main idea is to allow the representation of a sentence or word in document $A$ to be influenced by the representations of sentences and words in document $B$, and vice versa.

Consider the document vector $\mathbf{d}_A$ for document $A$.  In the HAN, the aggregation function considered only the (contextualized) vectors for sentences within the document (Equation~\ref{eq:aggregation}).  We inject another layer that ``attends'' to document $B$ and its sentences.  Let $\mathcal{B}$ denote the set containing all contextualized sentence vectors\footnote{Note that we have different strategies for different models here. Details are included in Appendix \ref{apx:sent_context}.} 
from $B$ and $B$'s document vector. We have: 
\begin{align}
    \mathbf{\tilde{d}}_A &= 
    \mathrm{affine} \left( \big[\mathbf{d}_A ; \sum_{\mathbf{v} \in \mathcal{B}}\frac{ \exp \mathbf{v}^\top \mathbf{d}_A}{\sum_{\mathbf{v'} \in \mathcal{B}} \exp \mathbf{v'}^\top \mathbf{d}_A} \mathbf{v} \big]
\right)
\label{eq:cda}
\end{align}
Again, we suppress the parameters of the affine transformation for clarity.  The new vector $\mathbf{\tilde{d}}_A$ is now used as the document representation for $A$. The same process is repeated in the other direction for obtaining the document vector $\mathbf{\tilde{d}}_B$ which is used as the document representation for the candidate document $B$.

The modification above is a variant of our model we call \textsc{Shallow}; it modifies only the final layer of the encoding so that $A$'s document vector depends on $B$'s document and sentence vectors, and vice versa. A similar layer can optionally be added to update each \emph{sentence} vector in $A$, using attention over the sentence and word vectors in $B$; this is illustrated in Figure~\ref{fig:model}, and we refer to it as the \textsc{Deep} variant of our model, because CDA is used to modify both sentence and document vectors.

More generally, CDA could be applied with additional levels in a HAN's hierarchy (e.g., paragraphs) and with different design choices about attention across levels.

\vspace{2mm}

\noindent \textbf{Relation to previous models}. Our idea is related to prior work which has encoded shorter texts such as sentences using attention over their syntactic structures \cite{liu18} to better align texts at the sentence-level. We go beyond sentences by encoding longer texts such as documents at multiple levels using attention over their document structures. An approach similar to ours is due to \citet{li19}, who used   cross-graph attention to compute alignment between computer programs.
However, they only evaluated document-to-document alignment (not other levels). 
They also rely on a graph representation of the documents, which may be costly both in terms of annotation and in computational cost for the required graph-based encoder; semantic and discourse graph structures for natural language are an interesting opportunity to explore in future work.

\section{A Benchmark for Document Relation Prediction and Localization }
\label{sec:benchmark}

While many tasks and datasets focus on understanding the relationships between sentences or documents separately, to the best of our knowledge, there are no joint publicly available English benchmark for both D2D and S2D tasks. Annotating document correspondences is expensive and time-consuming, especially at a fine-grained level like sentences. Therefore, we introduce a new benchmark consisting of six tasks (four D2D and two S2D).
This benchmark is shared publicly to encourage continued research.\footnote{Relevant details such as train/dev./test splits are included in  Appendix \ref{apx:benchmark-details}.}

\begin{table}[]
  \def\arraystretch{1.2}\tabcolsep=3.5pt
\begin{tabular}{lrrrrrr}
\toprule
    & \textbf{Pairs}& \textbf{Docs} &  \multicolumn{2}{c}{\textbf{Words}} &  \multicolumn{2}{c}{\textbf{Sentences}} \\  
  \textbf{Dataset}            & count & count &  avg & std &  avg & std  \\  \toprule
AAN         & 132K & 13K   & 122.7 & 11.2  &  4.9 & 2.7 \\ 
OC & 300K   & 567K   & 190.4 &16.3 &  7.0 & 3.5\\ 
S2ORC       & 190K    & 270K   & 263.7 & 19.2  &  9.3 & 5.9 \\
PAN         & 34K & 23K & 1569.7 & 90.4 & 47.4 & 66.1 \\
\toprule
\end{tabular}
\caption{Dataset statistics, namely the number of unique documents (count), average (avg) number of words per sentence and sentences per document along with their standard deviations (std).}
\smallskip
\label{table1}
\end{table}

\paragraph{Datasets.}
Our data resources of citation recommendation come from the ACL Anthology Network Corpus (AAN; \citealp{Radev2009TheAA}), the Semantic Scholar Open Corpus (OC; \citealp{Bhagavatula2018ContentBasedCR}), and the Semantic Scholar Open Research Corpus (S2ORC; \citealp{Lo2020S2ORCTS}). For plagiarism detection, we use the PAN plagiarism alignment task \citep{potthast:2013}.  We downsample OC and S2ORC, which are very large.
All of our datasets are preprossessed similarly: we filter out characters that are not digits, letters, punctuation, or white space in the texts. 

\paragraph{AAN.} Contains computational linguistics papers published on ACL Anthology from 2001 to 2014,  along with their metadata. For each paper, we extract its abstract and the abstracts of its citations and treat them as positive pairs without including full texts. 
For each positive pair's source paper, an (uncited, presumed irrelevant) negative paper is sampled at random to create a negative instance.  Since the dataset is not complete, we filter out pairs where either document lacks an abstract, but otherwise include all positive citation pairs. 
 
 \paragraph{OC.} Contains about 7.1M papers in computer science and neuroscience.  We follow a similar procedure to that for AAN.  Here we only select one citation per source paper, for wider coverage.  

  \paragraph{S2ORC.} A large contextual citation graph of  8.1M open access papers across broad domains of science. The papers in S2ORC are divided into sections and linked by citation edges. We select one section with at least one citation edge provided and the abstract of the cited paper to obtain a positive pair. To obtain a negative pair, we randomly select a paper from S2ORC which is not cited by the source section. Pairs with incomplete abstract or text are filtered out. To obtain the S2D ground-truth for a positive pair (which we will use only in evaluation, not as supervision), we use the citation span stored in the citation edge to identify where the citation appears in the citing document. Specifically, the information implied by the edges (that a paper cites another) for any pair is used to localize the ground-truth sentences which contain the citation.
 That citation in the sentence is removed from the text to prevent leakage, and the citing sentence is recorded. Note that not every pair has a citation edge that contains relevant sentence-level information, in which case the pair is discarded.

\vspace{3mm}
\noindent Note that for all the citation-related datasets above, the examples are counted as ``negative'' as long as they are uncited by the relevant paper. It is possible to use a heuristic approach to avoid treating similar documents as negative examples but we chose not to because the constraint is already largely satisfied with the random sampling procedure and the hypothesis that a paper should be cited by another one when they have high lexical overlap which may not always be true.

\paragraph{PAN.} A collection of web documents which contain several kinds of plagiarism phenomena. Human annotations show the segments of texts that are relevant to the plagiarism both in the source and suspicious documents. We construct a positive pair by extracting the relevant segment in the source document and a span (continuous) of text containing the relevant segment in the suspicious document. A negative pair is subsequently constructed by replacing the source segment in the afore-created positive pair with a segment from the corresponding source document which is not annotated as being plagiarised. For the S2D task, the sentences on the suspicious side that are not relevant to the plagiarism are treated as negative candidates in the positive pair. Note that the mapping between sentences is missing from the annotation, which prevents us from creating a sentence-to-sentence task.

\paragraph{Evaluation scores.} For D2D tasks, we report accuracy and $F_1$ score. For S2D tasks, we report the mean reciprocal rank (MRR) and precision-at-$N$ (P@$N$).  In the plagiarism case, multiple sentences can be seen as positive instances; MRR only considers the rank of the first relevant sentence, while P@$N$ reports the number of relevant sentences in the top $N$.

\section{Experiments} \label{sec:experiments}
Our experiments are performed on the above benchmark to test the benefit of cross-document attention. We first evaluate our model on scientific document citation recommendation (Section \ref{sec:citation_rec}) followed by citation localization (Section \ref{sec:citation_loc}). Then, we evaluate our model on web document plagiarism detection and localization (Section \ref{sec:plagiarism}). 

\subsection{Settings}

\vspace{3mm}

\paragraph{Baselines.}
We compare our method with previous established hierarchical document methods adapted for the task of similarity learning described in Section \ref{sec:han}. For baseline selection, we considered only methods that could deal with documents of arbitrary length on all of the examined datasets. In particular, we focus on two types of hierarchical attention networks (HANs), namely the first is using pretrained transformer representations from BERT and the other bidirectional GRU trained end-to-end:

\begin{itemize}
    \item \textbf{BERT-AVG}: represents each sentence with the average embedding of its tokens from BERT \cite{devlin19}. The representation of the document is computed as the average of the sentence representations. 
    
    \item \textbf{BERT-HAN}: uses BERT to represent sentences with the average embedding. Following \citet{Pappagari_2019}, the representation of the document is computed by the HAN network starting from the sentence-level representations. The model does not have direct access to word-level representations.
    
    \item \textbf{GRU-HAN}: encodes documents with a hierarchical attention network with word-level and sentence-level abstractions based on GRU \citep{yang16, jiang2019}. 
    
\end{itemize}

For our augmentation, we equip both types of hierarchical encoders with a \textsc{Shallow} or a \textsc{Deep} CDA component, keeping the base setup exactly the same. GRU and BERT are widely used contextualizers in NLP, while each can be viewed as a strong representative of the family of recurrent neural networks and transformers respectively. Note that BERT-HAN only trains a model over sentence-level representations, thus \textsc{Deep} does not apply to BERT-HAN. 
For the S2D task, we extract sentence representations from the candidate document $\mathbf{v} \in \mathcal{B}$ per model and rank them according to their similarity with the the target document vector $\mathbf{d}_A$ using an attention function:
\vspace{-2mm}
\begin{align}
\mathrm{AttScore}=\frac{ \exp \mathbf{v}^\top \mathbf{d}_A}{\sum_{\mathbf{v'} \in \mathcal{B}} \exp \mathbf{v'}^\top\mathbf{d}_A},
\end{align}
\noindent or a cosine similarity function:
\vspace{-2mm}
\begin{align}
\mathrm{CosScore}=\frac{\mathbf{v}^\top \mathbf{d}_A}{\|\mathbf{v}\|\|\mathbf{d}_A\|}.
\end{align}
We will  refer to them as \emph{attention alignment} and \emph{cosine alignment} respectively. The best scores for each metric and encoder type are marked in  \textbf{bold}. 

Note that the goal of our experiments was not to compare to state-of-the-art document models but to make a controlled experiment using various hierarchical configurations and provide some initial estimates of the difficulty of our benchmark for multilevel document alignment.

\begin{table*}[htp]
\centering
  \def\arraystretch{1.12}\tabcolsep=8pt    
\begin{tabular}{lcllllll|ll}
                      \toprule
                      &        & \multicolumn{2}{c}{\textbf{AAN}} & \multicolumn{2}{c}{\textbf{OC}} & \multicolumn{2}{c}{\textbf{S2ORC}} &  \multicolumn{2}{c}{\textbf{PAN}}\\
\textbf{Encoder}               & \textbf{CDA}  & Acc           & $F_1$       & Acc              & $F_1$           & Acc           & $F_1$      & Acc & $F_1$ \\ \toprule
  BERT-AVG & --    & 53.54       &  53.89  & 84.72                 & 84.99             &  77.78            &  76.92      & 79.62 & 76.60   \\ \hline
  \multirow{2}{*}{BERT-HAN}                    &  --   & 67.32       & 64.97         & 85.96                 & 86.33             &             90.67&  90.76   & \textbf{87.57} & \textbf{87.36}    \\
                      & \textsc{Shallow}  & \textbf{71.57}       &  \textbf{69.08}        & \textbf{87.81}                &  \textbf{87.89}            &              \textbf{91.92}&  \textbf{92.07}   & 86.23 & 86.19    \\  \hline
\multirow{3}{*}{GRU-HAN}  &  --   & 68.01       & 67.23   & 84.46           &   82.26           & 82.36        & 83.28 & 75.70 & 75.88 \\
                      & \textsc{Shallow} & 74.51       & 74.81   & 88.71           &  88.96            & 88.91        & 89.92  & \textbf{77.04} & \textbf{78.23} \\
                      & \textsc{Deep} & \textbf{75.08}       & \textbf{75.18}   & \textbf{89.79}           &  \textbf{89.92}            & \textbf{91.59}  & \textbf{91.61}   & 75.77 & 76.71    \\
                      \toprule
\end{tabular}
\caption{Comparison of our models with the HAN baseline using different encoders on document-to-document alignment over AAN, OC, and S2ORC datasets in terms of accuracy and $F_1$ score. }\smallskip
\label{table2}
\end{table*}

\vspace{3mm}

\paragraph{Configuration.} All the models are implemented in PyTorch. Our code is available on Github.\footnote{\url{https://github.com/XuhuiZhou/CDA}} We use Adam to optimize the parameters with an initial learning rate of  $10^{-5}$. The dimensions of hidden state vectors in GRUs and other hidden layers are set to 50 as in the original HAN \cite{yang16}. For word embeddings, we use 50-dimensional GloVe embeddings, which are updated during the training phase. 

\begin{table}[htp]
\centering
  \def\arraystretch{1.12}\tabcolsep=6pt    
\begin{tabular}{llll}
                      \toprule
\textbf{Encoder}               & \textbf{CDA}  & Acc           & $F_1$      \\ \toprule
\multirow{2}{*}{BERT-HAN} & --    & 73.36       & 73.51   \\
                      & \textsc{shallow} & \textbf{82.03} &  \textbf{82.08} \\  \bottomrule

\end{tabular}
\caption{Comparison with BERT-HAN using finetuning on document-to-document alignment on AAN.}\smallskip 
\label{table3}
\vspace{-3mm}
\end{table}
For pretrained contextualized embeddings, we use BERT-large implemented by HuggingFace.\footnote{\url{https://huggingface.co/transformers/model_doc/bert.html}} Note that, due to budget constraints, we keep BERT frozen in all experiments except for the finetuning experiment in Section~\ref{sec:citation_rec}. 
Unless otherwise noted, we perform early stopping based on the validation loss if there is no improvement for 5 consecutive epochs. 
The size of parameters of HAN models with GRU and BERT (kept frozen) are 20M and 1M respectively. 
Our corresponding models with a cross-document alignment component increase the number of parameters marginally, namely by 20K parameters. The networks with hierarchical configuration have $\mathcal{O}(T\log D )$ complexity with GRU and  $\mathcal{O}(T^2\log D )$ with BERT ($T$: sequence length, $D$: number of layers).  \textsc{Shallow} (\textsc{Deep}) adds one (two) more linear and quadratic terms respectively to these complexities, hence the asymptotic complexity remains the same. In practice, adding CDA negligibly impacts decoding speed.\footnote{See Appendix \ref{apx:computational cost} for details.} For more training details, see Appendix \ref{apx:setup}.

\subsection{Citation Recommendation} \label{sec:citation_rec}

We evaluate  the ability of our models to predict whether one document cites another, given 
citing signal at the document level, and specifically to quantify the effect of augmenting a model with CDA. 
From the results shown in Table \ref{table2} (left), we see a consistent benefit from CDA  across AAN, OC, and S2ORC, on accuracy and $F_1$.  Further, the \textsc{Deep} version of our model consistently outperforms the \textsc{Shallow} one on these tasks.

\begin{figure}[ht]
  \centering
  \includegraphics[width=\columnwidth,height=4cm]{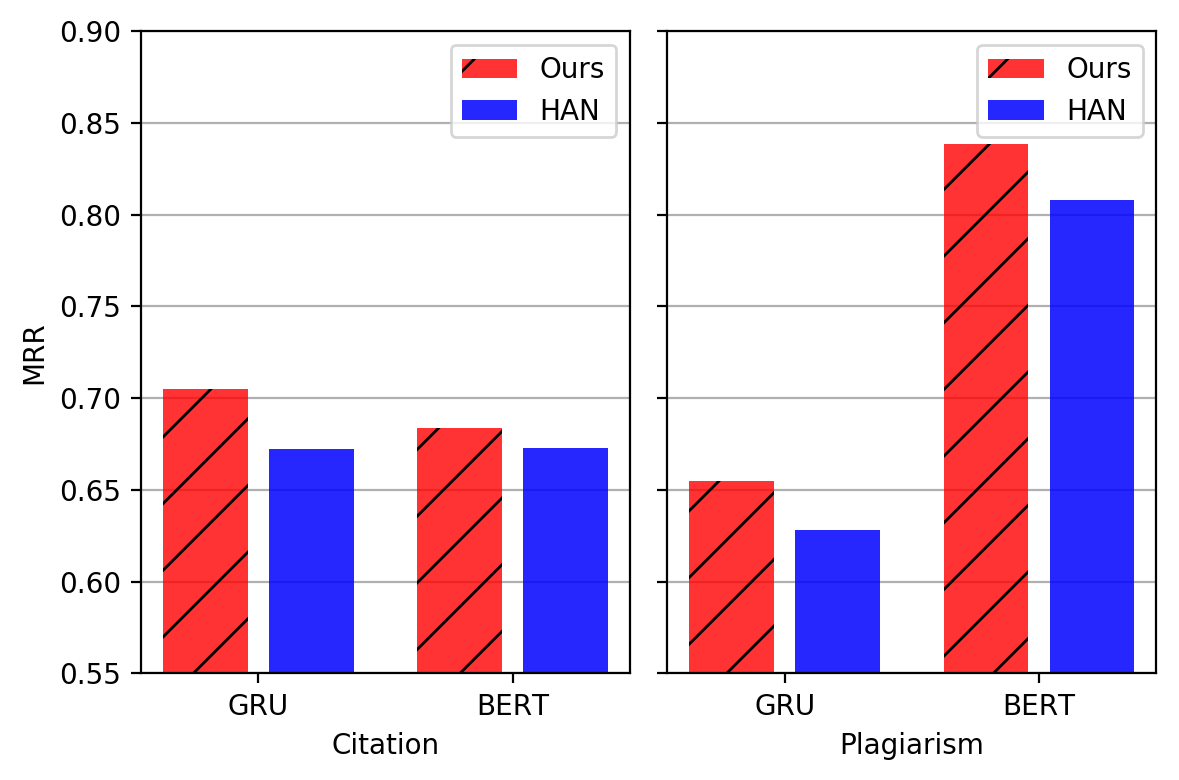}
  \vspace{-6mm}
  \caption{
  Citation localization results in terms of MRR given oracle document-to-document alignments. \textit{Left}: S2ORC. \textit{Right}: plagiarism detection. For GRU and BERT, ``ours'' refers to adding a \textsc{Deep} and \textsc{Shallow} CDA component, respectively. }
  \label{fig3}
  \vspace{-3mm}
\end{figure}
\paragraph{Finetuning BERT.}
To further evaluate our method, we finetune the BERT- HAN and \textsc{shallow} with BERT models, on the AAN dataset (where GRU models show an advantage).\footnote{Maximum 8 epochs, batch size 8; other hyperparameters set according to HuggingFace's GLUE task, \url{https://huggingface.co/transformers/examples.html}.} 
Table \ref{table2} shows that finetuning improves both models' performance, and the benefit of CDA is still present.

\paragraph{Comparison to state-of-the-art models.} 
our finetuned BERT-HAN with CDA (\textsc{Shallow})  is stronger than the SMASH model of \citet{jiang2019}, which achieves 80.68\% accuracy and 80.84\% $F_1$ on AAN. \citet{yang2020512} introduce a method similar to our BERT-HAN baseline and achieve 85.36\% accuracy and 85.43\% $F_1$. Note that both models carried out training on full texts; we only use abstracts, using a much smaller computational budget.
As reported in \citet{jiang2019}, the baseline HAN trained on full texts achieves 78.13\% accuracy, while HAN only achieves 68.01\% accuracy on our AAN task.

\begin{table*}[]
\centering
  \def\arraystretch{1.12}\tabcolsep=6.5pt    
\begin{tabular}{llllll|llll}
         \toprule
                        & & \multicolumn{4}{c|}{\textbf{Attention Alignment}} & \multicolumn{4}{c}{\textbf{Cosine Alignment}} \\ 
\textbf{Encoder}        & \textbf{CDA}       & MRR    & P@10   & P@5 &  P@1 & MRR    & P@10    & P@5  & P@1 \\ \toprule
    Random                    & --    & 0.3611  & 46.03  & 39.30  & 28.83 & 0.3611 & 46.03 & 39.30 &  28.83\\
                        \hline
   BERT-AVG & -- & 0.5596 & 72.67 & 64.24 & 47.22  & 0.5470 & 74.18 & 64.25 & 45.07\\
               \multirow{2}{*}{BERT-HAN}         & -- & 0.6068 & 82.75  & 70.67 &  51.20 & 0.5481 & 63.71   & 56.73 & 48.46\\ 
                        & \textsc{Shallow} & \textbf{0.6240} & \textbf{84.95}  & \textbf{73.14} & \textbf{52.51} & \textbf{0.6152}  & \textbf{81.21}  & \textbf{71.16} & \textbf{52.27} \\
                        \hline
\multirow{3}{*}{GRU-HAN}    & --    &0.5430  & 74.94   & 63.38  & 44.75 & 0.4742  & 52.59    & 47.91 & 41.68 \\
                        & \textsc{Shallow} &0.6225  &84.17    & 73.11  & 52.42 & 0.5013  & 54.19  & 48.93 & 45.31 \\
                        & \textsc{Deep} & \textbf{0.6474} & \textbf{86.37} & \textbf{76.11} & \textbf{54.90}  & \textbf{0.6252} & \textbf{81.93}     & \textbf{72.04}  & \textbf{53.35}\\

\toprule
\end{tabular}

\caption{Comparison of baselines and our models at the sentence-level S2ORC task. The \textit{Random} baseline assigns a random alignment score between 0 and 1 for each sentence. }\smallskip
\label{table4}
\end{table*}
\subsection{Citation Localization} \label{sec:citation_loc}

The same models as those in the D2D experiments above can be used to extract S2D alignments for evaluation on the second S2ORC task. If a model fails to predict the document alignment, the sentence-level alignment is counted as incorrect.

As shown in Table \ref{table3}, the deep variant of CDA here shows a consistent advantage over the shallow one, suggesting that explicitly modeling word-level correspondences helps localize citations. We also find that \emph{attention alignment} is consistently better than \emph{cosine alignment} in S2D tasks.

We also consider an oracle evaluation, where the trained models are given the correct D2D prediction (recall that above, the D2D and S2D alignments are jointly predicted).  Figure~\ref{fig3} (left) illustrates that CDA is beneficial in this setting as well, for both encoders. The other evaluation metrics show similar trends.

\subsection{Plagiarism Detection} \label{sec:plagiarism}

For plagiarism detection, the input consists of a source document and a suspicious document; the D2D task is to predict whether the suspicious document plagiarizes the source document, and the S2D (localization) task is to identify which sentences in the suspicious document plagiarize.
The dataset here (PAN) is considerably smaller than those we explored for citations (Table~\ref{table1}).

D2D results are shown in Table~\ref{table2} (right). CDA is not helpful to the BERT-HAN model and only \textsc{Shallow} CDA helps the HAN GRU model on the D2D task, which could be attributed to the small size of the plagiarism dataset.

S2D performance is shown in Table~\ref{plas2d} with \textit{attention alignment}; here we see a consistent benefit from CDA across encoders and evaluation scores. 

\begin{table}[]
\centering
  \def\arraystretch{1.2}\tabcolsep=3pt    
\begin{tabular}{llllll}
         \toprule
\textbf{Enc}        & \textbf{CDA}  &  MRR  & P@10 & P@5   \\ \toprule
        Random                & --  &  0.4215 & 44.23 & 43.28  \\
                        \hline
 BERT-AVG  & -- & 0.7864 & 58.24 & 64.69    \\
 \multirow{2}{*}{BERT-HAN}                       & -- & 0.8072 & 60.36   &68.94   \\ 
                        & \textsc{Shallow} & \textbf{0.8386} & \textbf{60.47} & \textbf{69.07}   \\
                        \hline
\multirow{3}{*}{GRU-HAN}    & --    &0.6205  & 50.72  & 51.90    \\
                        & \textsc{Shallow} & \textbf{0.6479}  & 51.71 & 53.05   \\
                        & \textsc{Deep} & 0.6378 & \textbf{52.07} & \textbf{53.82}   \\

\toprule
\end{tabular} 
\caption{Sentence-to-document plagiarism detection.}\smallskip 
\label{plas2d}
\vspace{-5mm}
\end{table}

\section{Other Related Work} 

\noindent \textbf{Latent Alignment.} \label{sec:latent_matching}
Attention has been previously used to align word sequences based on their intermediate hidden states for summarization \cite{rush15} and machine translation \cite{Bahdanau15}.  The alignment is typically softly learned and does not consider alternative alignments in a probabilistic sense. Hard attention \cite{luong15} is an alternative approach which selects only one word at a time but it is non-differentiable and requires more complicated techniques such as reinforcement learning to train.  \citet{NIPS2018_8179} considered an alternative attention network for learning latent variable alignment models based on amortized variational inference. 
Others modified attention to attend to partial segmentations and subtrees \cite{kim18} or trees \cite{liu18}, while  \citet{glomo18} cast the problem as latent graph learning to capture dependencies between pairs of words from unlabeled data.  Orthogonal to these studies, we use attention to compare documents represented by hierarchical document encoders at multiple  levels. 

\vspace{4mm}
\noindent \textbf{Similarity Learning.} 
There are three types of similarity learning in NLP. The supervised paradigm differs from typical supervised learning in that training examples are cast into pairwise constraints \cite{yang2006distance}, as in cross-lingual word embedding learning based on word-level alignments \cite{faruqui14}  and zero-shot utterance/document classification \cite{yazhen15,nam16,pappas19b} based on utterance/document-level annotations. The unsupervised paradigm aims to learn an underlying low-dimensional space where the relationships between most of the observed data are preserved, as in word embedding learning  \cite{collobert11,word2vec,pennington-etal-2014-glove,NIPS2014_5477}. The weakly supervised paradigm is the middle ground between the two, as in cross-lingual word embedding learning based on sentence-level alignments \cite{hermann14,gows15}. Our approach is  weakly supervised and operates at the document-level, making use of structural correspondence between documents.

\section{Conclusion}

We augment hierarchical attention networks with cross-document attention, allowing their use in document-to-document and sentence-to-document alignment tasks.  
We introduce benchmarks, based on existing datasets, to evaluate model performance on such tasks.
In controlled experiments, we observe a benefit from cross-document attention on three out of the four document-to-document tasks and two out of two sentence-to-document tasks.

\section*{Acknowledgments}
The authors thank Efstathios Stamatatos, Phillip Keung, Lucy Lin, and the anonymous reviewers for their helpful feedback. Nikolaos Pappas was supported by the Swiss National Science Foundation under the project UNISON, grant number P400P2\_183911.  This work was supported in part by US NSF grant 1562364.

\bibliography{references_v2}
\bibliographystyle{acl_natbib}

\clearpage
\appendix

\section{Supplementary Material for ``Multilevel Text Alignment with Cross-Document Attention''}
 We provide more details about the datasets in the new benchmark, experimental setup, choices of hyperparameters, model design choices, and validation performances corresponding to the ones reported in the main paper. Moreover, we provide qualitative examples that visualize the cross-document attention component.

\subsection{Benchmark Details}\label{apx:benchmark-details}
In Table \ref{splits}, we list the statistics about training, development and test splits of the four datasets that are part of our benchmark. The exact splits used in our experiments are released along with the datasets of the benchmark.\footnote{\url{ https://xuhuizhou.github.io/Multilevel-Text-Alignment/}} A sample of our datasets are attached along with our submission. Note that we have not included the whole benchmark because its size exceeds the limit allowed in the submission portal. 

\begin{table}[htp]
\centering
  \def\arraystretch{1.2}\tabcolsep=7.5pt    
\begin{tabular}{cccc}
\toprule
  \textbf{Dataset}  & \textbf{Training}& \textbf{Validation} & \textbf{Test} \\
  \toprule
AAN         & 106,592 & 13,324   & 13,324  \\ 
OC          & 240,000 & 30,000  & 30,000 \\ 
S2ORC       & 152,000   & 19000   & 19000  \\
PAN         & 17,968 & 2,908 & 2,906 \\
\toprule
\end{tabular}
\caption{Dataset statistics regarding the number of examples for the  training/validation/test splits.} 
\label{splits}
\end{table}

\subsection{Experimental Setup Details}
\label{apx:setup}
For our experiments, we used the following computing infrastructure: 1 GeForce 960, 1 GeForce 1080, and 1 Titan Xp for the model training. The batch size is set to 128 for GRU-HAN (including our augmentation) experiments on AAN task, and is set to 256 for all other experiments. The running time ranges from 36 hours to 48 hours for the GRU-based models, and from 1 to 2 hours for BERT-frozen models, and about 24 hours for BERT-finetuning models.

\subsection{Development Scores}
We report validation performance for all the  reported test results for the document-to-document alignment tasks in Tables \ref{core3}--\ref{core2}. Note that for the sentence-to-sentence alignment tasks there is no validation taking place, hence, there are no development scores to report here.

\vspace{3mm}

\subsection{Model Design Choices}
In this section, we describe the set of model design choices that were made based on development performance before running our main experiment.

\subsubsection{Word Embedding Dimension}
To decide what embedding size to use for our main experiments, we experimented with 50-dimensional and 200-dimensional GloVe embeddings by training the GRU-HAN model on the AAN task. When keeping other settings exactly the same as the aforementioned GRU-HAN models on the AAN task, the performance of GRU-HAN with a larger word embedding size is lower than the 50-dimensional model as shown in Table \ref{word2vec_dim}. Therefore, we stick with 50-dimensional GloVe embeddings for the other experiments. 

\begin{table}[ht]
\centering
\begin{tabular}{llll}
                      \toprule
\textbf{Encoder}               & \textbf{Dim}  & Acc           & $F_1$      \\ \toprule
\multirow{2}{*}{HAN} & 50    & 68.01       & 67.23  \\
                      & 200 & 66.94 & 66.24  \\  \bottomrule

\end{tabular}
\caption{Influence of the dimensionality of word embeddings to the baseline model HAN. }\smallskip 
\label{word2vec_dim}
\vspace{-3mm}
\end{table}

\subsubsection{Sentence Contextualization}
\label{apx:sent_context}
Preliminary experiments show that one could obtain better performance on the AAN D2D task by using sentence vectors before contextualization in Equation \ref{eq:cda} for GRU-based models. Therefore, the experiments for GRU-based models above use sentence vectors before contextualization for CDA. For BERT-based models, we use two GRU layers to contextualize sentence vectors, the sentence vectors after the first GRU layer are used in CDA. Practitioners can be flexible in deciding how CDA is used for different tasks and encoders.

\begin{table}[ht]
\centering
\begin{tabular}{llll}
                      \toprule
\textbf{Encoder}               & \textbf{CDA}  & Acc           & $F_1$      \\ \toprule
\multirow{2}{*}{BERT-HAN} & --    & 75.41       & 74.25   \\
                      & \textsc{shallow} & \textbf{83.72} &  \textbf{82.57} \\  \bottomrule

\end{tabular}
\caption{Development set results corresponding to Table \ref{table3}.}\smallskip 
\label{core3}
\vspace{-3mm}
\end{table}

\begin{table*}
\centering
  \def\arraystretch{1.15}\tabcolsep=8pt  
\begin{tabular}{lcllllll|ll}
                      \toprule
                      &        & \multicolumn{2}{c}{\textbf{AAN}} & \multicolumn{2}{c}{\textbf{OC}} & \multicolumn{2}{c}{\textbf{S2ORC}} &  \multicolumn{2}{c}{\textbf{PAN}}\\
\textbf{Encoder}               & \textbf{CDA}  & Acc           & $F_1$       & Acc              & $F_1$           & Acc           & $F_1$      & Acc & $F_1$ \\ \toprule
  BERT-AVG & --    & 54.72       &  54.12  & 84.58                 & 84.67             &  80.52            &  81.14      & 82.73 & 82.42   \\ \hline
  \multirow{2}{*}{BERT-HAN}                    &  --   & 67.34       & 64.69        & 85.73                 & 86.23             &             90.46&  90.49   & \textbf{88.85} & \textbf{88.77}    \\
                      & \textsc{Shallow}  & \textbf{73.20}       &  \textbf{70.80}        & \textbf{87.73}                &  \textbf{87.91}            &              \textbf{91.66}&  \textbf{91.54}   & 86.76 & 86.79    \\  \hline
\multirow{3}{*}{GRU-HAN}  &  --   & 69.68       & 69.15   & 83.06           &   83.84           & 82.65        & 83.41 & 76.40 & 76.77 \\
                      & \textsc{Shallow} & 77.46       & 75.41   & 89.31           &  89.41            & 89.78        & 89.89  & \textbf{77.02} & \textbf{78.15} \\
                      & \textsc{Deep} & \textbf{78.17}       & \textbf{75.94}   & \textbf{91.10}           &  \textbf{91.11}            & \textbf{92.01}  & \textbf{92.02}   & 76.99 & 78.01    \\
                      \toprule
\end{tabular}
\vspace{-2mm}
\caption{Development set results corresponding to Table \ref{table2}. 
}\smallskip
\vspace{-5mm}
\label{core2}
\end{table*}

\subsection{Integrating Cross-Document Attention}
For the integration of the cross-document attention representations with the representations of the hierarchical attention network we experimented with two  options, namely concatenation and addition of vectors. We found that our method is more competitive with concatenation. However,  integrating with  concatenation involves a linear projection to match the original hidden size of the network which increases slightly the number of parameters. Here, we evaluate the performance of our model when it uses addition, that is when the number of parameters remains exactly the same with that of the base network. 

We evaluated the performance of our \textsc{Shallow} augmentation on BERT-HAN with finetuning BERT end-to-end. The results are displayed in Table \ref{addition}. With \textsc{Shallow} (addition), BERT-HAN achieves 79.02\% accuracy and 79.08\% $F_1$, which still improves over the BERT-HAN baseline. Interestingly, using addition instead of concatenation performs quite well and its performance is still better than the hierarchical attention network baseline. Hence, we conclude that the additional number of parameters is not the only factor responsible for the superior performance of our model.  

\begin{table}[htp]
\centering 
  \def\arraystretch{1.1}\tabcolsep=6pt    
\begin{tabular}{lccc}
                      \toprule
\textbf{Encoder}               & \textbf{CDA}  & Acc           & $F_1$      \\ \toprule
\multirow{5}{*}{BERT-HAN} & --    & 73.36       & 73.51   \\
                      & \textsc{shallow}& \multirow{2}{*}{\textbf{82.03}} &   \multirow{2}{*}{\textbf{82.08}} \\  
                      &  (concatenation) &  &    \\ 
                      & \textsc{shallow} & \multirow{2}{*}{{79.02}} &  \multirow{2}{*}{{79.08}} \\
                      &  (addition) &  &    \\ 
                      \bottomrule

\end{tabular}
\caption{Comparison with BERT-HAN using finetuning on document-to-document alignment on AAN.}\smallskip 
\label{addition}
\vspace{-3mm}
\end{table}
\subsection{Qualitative Inspection}
\begin{figure*}[h]
    \centering
        \includegraphics[clip, trim=0cm 14cm 0cm 0cm, width=1.00\textwidth]{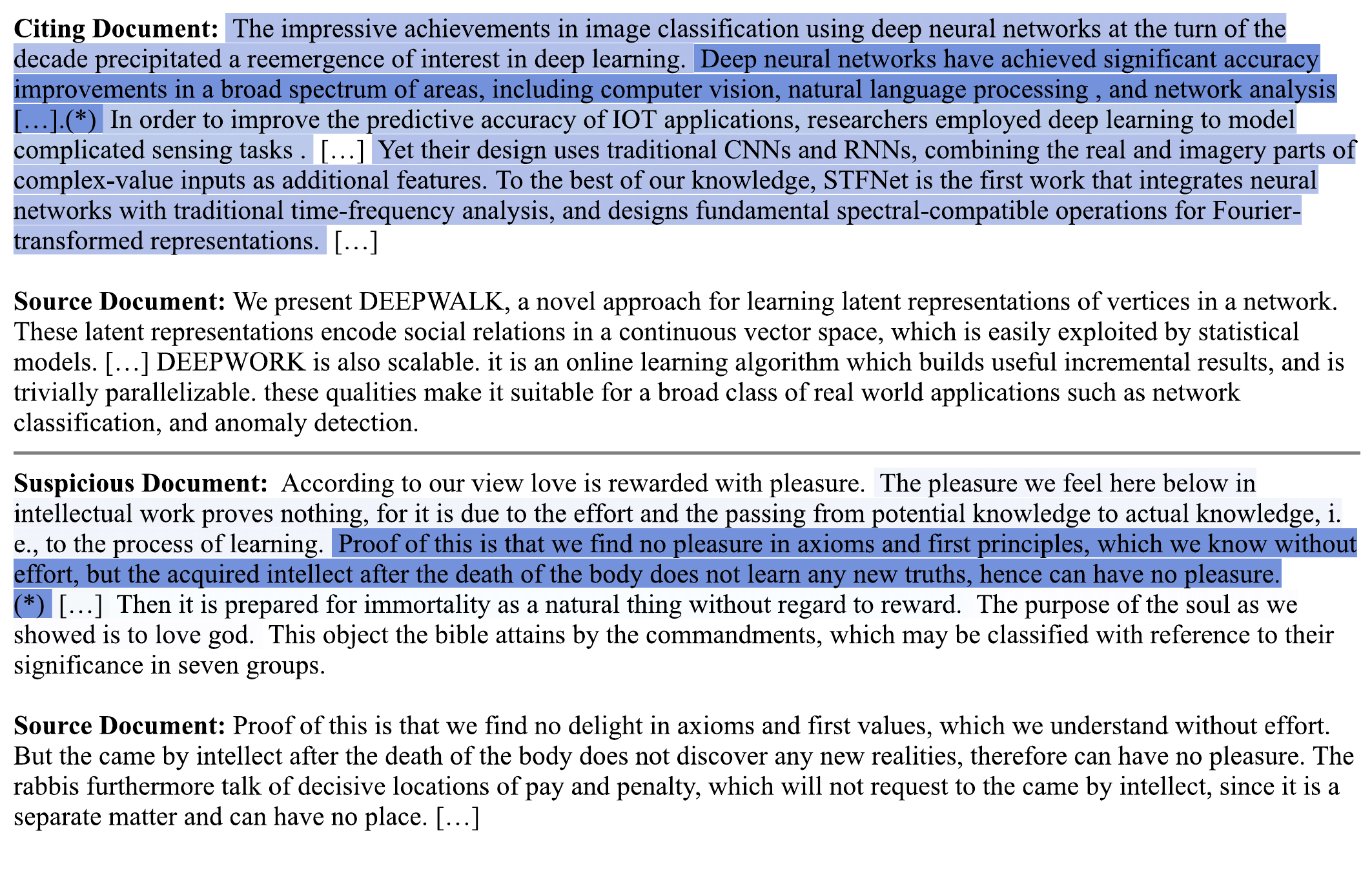}
    \caption{Example of BERT HAN \textsc{shallow}'s prediction on citation recommendation (above) and plagiarism detection (below). The attention scores produced for each sentence by HAN \textsc{shallow} are represented by the degree of blueness. The positive sentence is marked with an asterisk at the end. }
    \label{fig:examples}
\end{figure*}

We select two examples from the test sets of GORC and PAN, where both HAN and HAN \textsc{shallow} with BERT obtain correct D2D results. However, while HAN is confused of finding the sentence where citation or plagiarism happens, HAN \textsc{shallow} is able to locate the relevant sentences in the document as shown in Figure \ref{fig:examples}.

We find that the document-to-document results in these two examples are heavily dependent on the localization of the sentences. While we have difficulty in interpreting HAN's decision for the two examples, it is not hard for us to see how HAN \textsc{shallow}, as a unified model for D2D and S2D tasks, obtains its decision on whether this document cites or plagiarizes the other one. This property should be important for future models to pursue instead of simply producing a yes or no decision. 

\subsection{Computational Cost of CDA}
\label{apx:computational cost}
In Table \ref{computcost}, We show the average inference time of each epoch (256 batch size) on GeForce 1080 for tasks S2ORC and PAN, which have longer texts among our tasks. In general, the extra computational cost for CDA is negligible (1--2\% extra wall time), especially for \textsc{shallow}. Note that BERT-based models share the same property.

\begin{table}[htp]
\centering 
  \def\arraystretch{1.1}\tabcolsep=6pt    
\begin{tabular}{lccc}
                      \toprule
\textbf{Encoder}               & \textbf{CDA}  & S2ORC           & PAN    \\ \toprule
\multirow{3}{*}{HAN} & --    & 0.256      & 0.568   \\
                      & \textsc{shallow}& 0.256 & 0.581   \\  
                      
                      & \textsc{deep} & 0.258  & 0.637 \\
                   
                      \bottomrule

\end{tabular}
\caption{Comparison of average inference time (s) of each epoch for S2ORC and PAN.}\smallskip 
\label{computcost}
\vspace{-3mm}
\end{table}

\end{document}